\documentclass{article}

\usepackage{microtype}
\usepackage{graphicx}
\usepackage{subfigure}
\usepackage{booktabs} % for professional tables

\usepackage{hyperref}

\usepackage{amssymb}
\usepackage{amsbsy}
\usepackage{amsmath}
\def\bx{\mathbf{x}}
\def\bX{\mathbf{X}}
\def\bz{\mathbf{z}}
\def\bZ{\mathbf{Z}}
\usepackage{amsthm}
\theoremstyle{plain}

\theoremstyle{definition}
\usepackage[accepted]{icml2020}

\icmltitlerunning{Robust Variational Autoencoder for Tabular Data with $\beta$ Divergence}

\begin{document}

\twocolumn[
\icmltitle{Robust Variational Autoencoder for Tabular Data with $\beta$ Divergence}




\begin{icmlauthorlist}

\icmlauthor{Haleh Akrami}{usc}
\icmlauthor{Sergul Aydore}{aws}
\icmlauthor{Richard M. Leahy}{usc}
\icmlauthor{Anand A. Joshi}{usc}
\end{icmlauthorlist}

\icmlaffiliation{aws}{Amazon Web Service, New York, USA}
\icmlaffiliation{usc}{Signal and Image Processing Institute, University of Southern California, USA}

\icmlcorrespondingauthor{Sergul Aydore}{sergulaydore@gmail.com}
\icmlcorrespondingauthor{Haleh Akrami}{akrami@usc.edu}

\icmlkeywords{Machine Learning, ICML}

\vskip 0.3in
]



\printAffiliationsAndNotice{}  % leave blank if no need to mention equal contribution

\begin{abstract}
We propose a robust variational autoencoder with $\beta$ divergence for tabular data (RTVAE) with mixed categorical and continuous features. Variational autoencoders (VAE) and their variations are popular frameworks for anomaly detection problems. The primary assumption is that we can learn representations for normal patterns via VAEs and any deviation from that can indicate anomalies. However, the training data itself can contain outliers. The source of outliers in training data include the data collection process itself (random noise) or a malicious attacker (data poisoning) who may target to degrade the performance of the machine learning model. In either case, these outliers can disproportionately affect the training process of VAEs and may lead to wrong conclusions about what the normal behavior is. In this work, we derive a novel form of a variational autoencoder for tabular data sets with categorical and continuous features that is robust to outliers in training data. Our results on the anomaly detection application for network traffic datasets demonstrate the effectiveness of our approach.
\end{abstract}

\section{Introduction}
\label{sec:introduction}
An anomaly is defined as an observation that does not conform to normal patterns in the data. Early detection of anomalies is crucial for decision making systems to ensure undisrupted business. Anomaly detection is used in a wide variety of applications such as fraud and intrusion detection, military surveillance and medical diagnosis \cite{chandola2007outlier}. Our motivating application in this work is detecting malicious activities from network traffic data that compromise both categorical and continous features.

Because of limited labeled data, unsupervised machine learning algorithms such as one-class SVM \cite{erfani2016high}, K-Means \cite{munz2007traffic}, principal component analysis (PCA) \cite{chandola2007outlier} and variational autoencoders (VAEs) \cite{an2015variational,yao2019unsupervised} are are frequently adopted for anomaly detection problems. These methods are based on intrinsic properties of the dataset and do not require any labels.Their core idea is to learn the representations in the original or some latent feature space and then detect anomalies by computing the deviation from normal patterns. Among these methods, spectral anomaly detection approaches such as PCA and autoencoders try to find lower dimensional representations of the original data\cite{an2015variational}.Based on the assumption that anomalies and normal data are separable in this low dimensional representation. Once lower dimensional representations are learned, the data is reconstructed back in the original dimension. Reconstruction error between the original data and reconstructed data is used as a score to detect anomalies.Autoencoders and their variations are the core of most deep-learning based unsupervised anomaly detection. 

VAEs \cite{kingma2013auto} are generative models that adopt variational inference and graphical models. The advantage of the VAE over PCA and autoencoders is that it can learn the distribution of the data which provides a reconstruction probability in addition to reconstruction error as the anomaly score \cite{an2015variational}. VAE has two components: an encoder and a decoder. The encoder transforms high dimensional data to a low-dimensional latent space with an approximate tractable posterior distribution. The decoder samples from this distribution and transforms the sample back to the original dimension. The VAE minimizes two terms: the reconstruction cost and the regularizer. The regularizer penalizes any discrepancy between the prior distribution of the latent representations and the distribution induced by the encoder. 

The main assumption behind the use of latent space representations is that the training data is clean and represents normal behavior. However, in practice, training data can inevitably contain outliers or anomalies.The presence of outliers can have a disproportionate impact on training due to large negative log-likelihood values. Hence, not only does the model's representation of normal behavior degrade but also it may treat outliers as normal samples during inference. Therefore, achieving robustness to outliers is crucial in unsupervised models for accurate detection of outliers.

\subsection{Our Contribution}
Our work can be viewed as an extension of \citet{akrami2019robust} to tabular data. In order to achieve robustness to outliers, previous approaches
focus on modification of network architectures, adding constraints or modeling of outlier distribution \cite{zhai2017robust,eduardo2019robust}. In contrast,  \citet{akrami2019robust}  adopt $\beta$-divergence from robust statistics \cite{futami2017variational}. The log-likelihood term that VAE uses in the reconstruction loss minimizes the KL-divergence between the empirical distribution of the data and the parametric distribution at the output of the decoder. After demonstrating the non-robustness of the KL divergence, \citet{akrami2019robust} replaced the KL-divergence for data fitting with a robust $\beta$-divergence \cite{basu1998robust}. However, their derivations and implementations were limited to images. In this work, we derive a formulation for categorical variables and propose a training mechanism for tabular datasets.We show that the proposed approach works more accurately than the standard VAE using the publicly available tabular network traffic datasets.

\section{Variational Autoencoders}
\label{sec:variational_autoencoders}
In this section, we provide a review of VAEs. We adopt the notation in \citet{ghosh2019variational}. Let $\mathcal{X} = \left\{ \bx_i \right\}_{i=1}^N$ be high-dimensional i.i.d. samples drawn from the true data distribution $p_{\textrm{data}}(\bx)$ over a random variable $\bX$. Generative modeling aims to learn a mechanism from $\mathcal{X}$ to draw new samples such that $\bx_{\textrm{new}} \sim p_{\textrm{data}}$. VAEs provide a framework to achieve this goal by learning a representation in low-dimensional latent space. The generative process of the VAE is defined as
\begin{equation}
\bz_{\textrm{new}} \sim p(\bZ) \quad \quad \quad \bx_{\textrm{new}} \sim p_{\theta}(\bX | \bZ = \bz_{\textrm{new}})
\end{equation}
where $p(\bZ)$ is a fixed prior distribution over latent space $\bZ$. A stochastic decoder
\begin{equation}
D_{\theta}(\bz) = \bx \sim p_{\theta}(\bx \mid \bz) = p(\bX | g_{\theta}(\bz))
\end{equation}
maps the latent variable to the input space via the \textit{likelihood} distribution $p_{\theta}$, where $g_{\theta}$ is a non-linear function, typically a neural network, parameterized by $\theta$. Consequently, a VAE estimates $p_{\textrm{data}}(\bx)$ as the infinite mixture model $p_{\theta}(\bx) = \int p_{\theta}(\bx \mid \bz) p(\bz) d\bz$. At the same time, the input space is mapped to the latent space via a stochastic encoder
\begin{equation}
E_{\phi}(\bx) = \bz \sim q_{\phi}(\bz \mid \bx) = q(\bZ \mid f_{\phi}(\bx))
\end{equation}
where $q_{\phi}(\bz \mid \bx)$ is the \textit{posterior} distribution given by another non-linear function $f_{\phi}$ parameterized by $\phi$.

Computing the marginal log-likelihood $\log p_{\theta}(\bx)$ is generally intractable. Therefore, it is common to follow a variational approach which focuses on maximizing the evidence lower bound (ELBO) for a sample  $\bx$:
\begin{equation}
\begin{split}
\log  &p_{\theta}(\bx) \geq  \textrm{ELBO}(\phi, \theta, \bx) \\
&= \mathbb{E}_{\bz \sim q_{\phi}(\bz \mid \bx)} \log p_{\theta}(\bx \mid \bz) - \mathbb{KL}(q_{\phi}(\bz \mid \bx) || p(\bz)) \\
&\triangleq -\mathcal{L}_{\textrm{ELBO}}
\end{split}
\label{eqn:elbo}
\end{equation}
Maximizing RHS of equation \ref{eqn:elbo} over data $\mathcal{X}$ with respect to parameters $\phi$ and $\theta$ corresponds to minimizing the loss
\begin{eqnarray}
\arg \min_{\phi, \theta} && \mathbb{E}_{\bx \sim p_{\textrm{data}}}  \mathcal{L}_{\textrm{ELBO}} \\
&=& \mathbb{E}_{\bx \sim p_{\textrm{data}}} \left[ \mathcal{L}_{\textrm{REC}} + \mathcal{L}_{\textrm{KL}} \right]
\end{eqnarray}
where $\mathcal{L}_{\textrm{REC}}$ and $\mathcal{L}_{\textrm{KL}}$ are defined for sample $\bx$ as follows:
\begin{eqnarray}
\mathcal{L}_{\textrm{REC}} &=& - \mathbb{E}_{\bz \sim q_{\phi}(\bz \mid \bx)} \log p_{\theta}(\bx \mid \bz) \\
\mathcal{L}_{\textrm{KL}} &=& \mathbb{KL}(q_{\phi}(\bz \mid \bx) || p(\bz)).
\label{eqn:L_rec_kl}
\end{eqnarray}
The reconstruction loss $\mathcal{L}_{\textrm{REC}}$ computes the quality of encoded samples $\bx$ through $D_{\theta} \left( E_{\phi}(\bx) \right)$. The KL-divergence term $\mathcal{L}_{\textrm{KL}}$ measures the similarity between $q_{\phi}(\bz \mid \bx)$ and the prior $p(\bz)$ for each $\bz$. This KL-divergence term is also called the regularizer term since it acts as a regularizer during training \cite{hoffman2016elbo}.

\section{Robust Variational Inference}
In this section, we show how the reconstruction term $\mathcal{L}_{\textrm{REC}}$ can be modified using a robust divergence in order to make it more robust to outliers for categorical data. This approach was first proposed in \cite{akrami2019robust} for Gaussian and Bernoulli variables. Here, we will provide an extension for categorical  variables. 

Let the empirical distribution of $\bX$ be
\begin{equation}
\hat{p}(\bX) = \frac{1}{N} \sum_{i=1}^N \delta(\bX, \bx_i)
\end{equation}
where $\delta$ is the Dirac delta function. The KL-divergence between this empirical distribution $\hat{p}(\bX)$ and $p_{\theta}(\bX | \bz)$ can be written as
\begin{equation}
\begin{split}
 & \mathbb{KL}\left(\hat{p}(\bX) || p_{\theta}(\bX | \bz)  \right) = \int \hat{p}(\bX) \log \frac{\hat{p}(\bX)}{p_{\theta}(\bX | \bz)} d \bX  \\
& \quad \quad = \textrm{const} - \int \hat{p}(\bX) \log p_{\theta} (\bX \mid \bz) d \bX  \\
& \quad \quad = \textrm{const} - \int \frac{1}{N} \sum_{i=1}^N \delta(\bX, \bx_i) \log p_{\theta}(\bX \mid \bz) d \bX \\
& \quad \quad = \textrm{const} - \frac{1}{N} \sum_{i=1}^N \log p_{\theta}(\bx_i \mid \bz)
\end{split}
\end{equation}
indicating that maximizing the log-likelihood of a sample $\bx_i$ is equivalent to minimizing KL-divergence between the empirical distribution and the generative distribution for one sample. Let the KL-divergence for a single sample $\bx_i$ be
\begin{equation}
\mathbb{KL}^i \triangleq -\frac{1}{N} \log p(\bx_i \mid \bz).
\end{equation}
Then, the reconstruction loss for a single sample can be written as
\begin{equation}
\mathcal{L}_{\textrm{REC}}^i = N \mathbb{E}_{\bz \sim q_{\phi}(\bz \mid \bx)} \left [ \mathbb{KL}^i \right ].
\end{equation}
The log-likelihood term $ \log p(\bx_i \mid \bz)$ in $\mathbb{KL}^i$ is sensitive to the outliers because the negative log-likelihood of low probability samples can be arbitrarily high. Rather than using KL-divergence, it is possible to choose a different divergence measure to quantify the similarity between $\hat{p}(\bX)$ and $p_{\theta} (\bX \mid \bz)$. We use $\beta$-divergence which is defined as
\begin{equation}
\begin{split}
\mathbb{D}_{\beta}(\hat{p}(\bX) ||  p_{\theta} (\bX & \mid \bz) = \\
 & \frac{1}{\beta} \int \hat{p}(\bX)^{\beta + 1} d \bX \\
&  - \frac{\beta + 1}{\beta} \int \hat{p}(\bX) p_{\theta}(\bX \mid \bz)^{\beta} d \bX \\
&  + \int p_{\theta}(\bX \mid \bz)^{\beta + 1} d \bX
\end{split}
\end{equation}
which converges to $\mathbb{KL}$ as $\beta \rightarrow 0$.  It can be shown that minimizing $\beta$-divergence is equivalent to minimizing $\beta$-cross-entropy \cite{eguchi2010entropy, futami2017variational} which is defined as
\begin{equation}
\begin{split}
\mathbb{H}_{\beta} & (\hat{p}(\bX) || p_{\theta}(\bX | \bz)) = \\
& - \frac{\beta + 1}{\beta} \int \hat{p}(\bX) \left(p_{\theta}(\bX | \bz)^{\beta}-1 \right) d \bX  \\ 
& \quad + \int p_{\theta}(\bX| \bz)^{\beta + 1} d \bX.
\end{split}
\label{eqn:beta_entropy}
\end{equation}

\begin{algorithm}[!b]
   \caption{Training RTVAE}
   
   \textbf{Input:} 
   
	\hspace{\parindent} Initialize the parameters of the encoder $q_{\phi}(\bz \mid \bx)$ and the decoder $p_{\theta}(\bx \mid \bz)$.
		
	\hspace{\parindent} Robust divergence coefficient $\beta \geq 0$.

   \textbf{Output:} 
	\hspace{\parindent} $\phi$, $\theta$

\begin{algorithmic}[1]
   \WHILE{$\phi$ and $\theta$ not converged}
      \STATE Initialize $\mathcal{L}_{\textrm{REC-}\beta}=0 $
   \STATE Sample $\{\bx_1, \cdots, \bx_N \}$ from the training set
   \STATE Sample $\{\bz_1, \cdots, \bz_N \}$ from the prior $p(\bz)$
   \STATE Sample $\{\tilde{\bz}_1, \cdots, \tilde{\bz}_N \}$ from $q_{\phi}(\bz \mid \bx_i)$
   \FOR {$j$ in features}
   \STATE Compute $\beta$-divergence term: 
	\IF   {$j$ is a categorical variable}
   \STATE $\mathcal{L}_{\textrm{REC-}\beta} += - \frac{\beta+1}{N \beta} \sum_{i=1}^N \left( p_{\theta}(\bx_i^j \mid \tilde{\bz}_i)^{\beta} - 1 \right)$ \\  \quad \quad  \quad \quad \quad $+ \sum_{k=1}^K p_{\theta}(\bX^j == k \mid \tilde{\bz}_i)^{\beta+1}$
   \ELSIF {$j$ is a continous variable}
   \STATE $\mathcal{L}_{\textrm{REC-}\beta} +=$ \\
    \quad $- \frac{\beta+1}{N \beta} \sum_{i=1}^N \frac{1}{(2 \pi \sigma^2)^{\beta/2}} \exp \left( - \frac{\beta}{2 \sigma^2} \| \hat{\bx}_i^j - \bx_i^j \|^2 \right)$ \\ \quad $+\frac{\beta + 1}{\beta}$
   \ENDIF
   \ENDFOR
   \STATE Compute the regularizer : \\\quad $\mathcal{L}_{\textrm{KL}} = \mathbb{KL}({q_{\phi}(\bz \mid \bx_i) || p(\bz)}) $
   \STATE Update $\phi$ and $\theta$ by descending the total loss \\\quad $\mathcal{L}_{\textrm{TOT}} = \mathcal{L}_{\textrm{REC-}\beta} + \mathcal{L}_{\textrm{KL}} $
   \ENDWHILE
   \STATE Return $\phi$, $\theta$
\end{algorithmic}
\label{alg:rvae}
\end{algorithm}

 \begin{figure*}[t!]
	\centering
	\includegraphics[scale=.33]{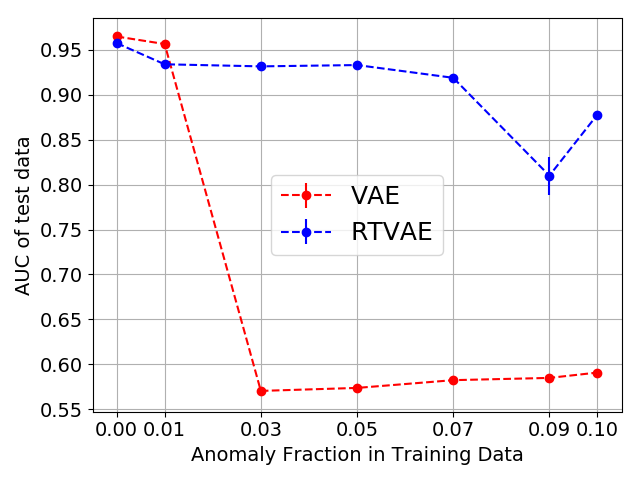}
	\includegraphics[scale=.33]{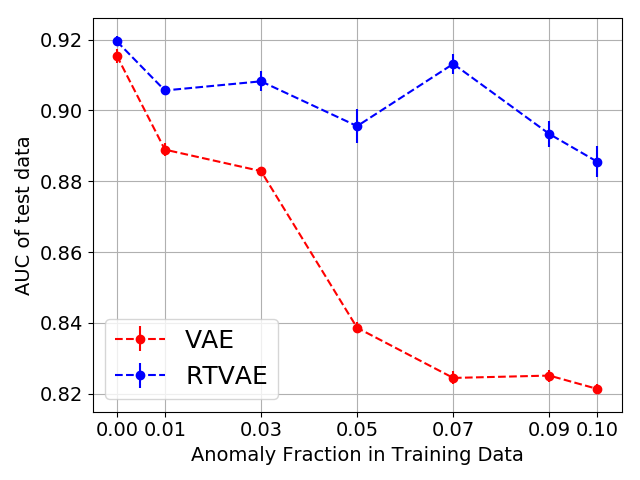}
	\includegraphics[scale=.33]{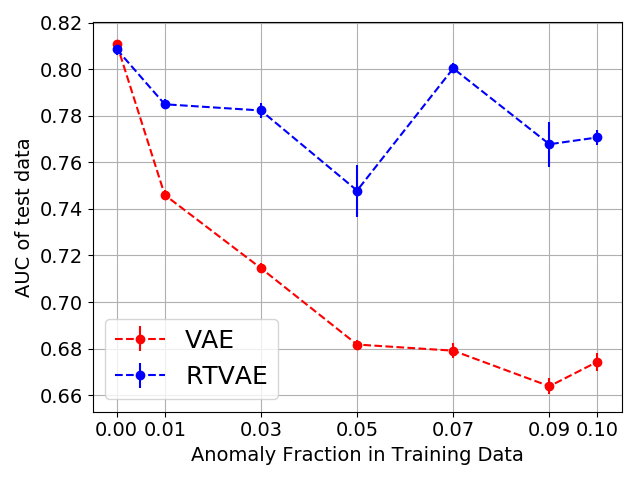}
	\caption{Performance comparison of VAE and RTVAE as a function of contamination in training data for datasets KDDCup99 (left), NSLKDD (middle), and UNSW-NB15 (right)}.
	\vspace{-1em}
	\label{fig:results}
\end{figure*}

Since we are interested in applying VAE to a categorical data, we can assume  that the generative distribution is a categorical distribution with $K$ categories. Then, the first integral in equation \ref{eqn:beta_entropy} becomes:
\begin{equation}
\begin{split}
\int \hat{p}(\bX) &  p_{\theta}(\bX | \bz)^{\beta }d \bX = \\
& \frac{1}{N} \int \sum_{i=1}^N \delta(\bX, \bx_i) \left( p_{\theta}(\bX | \bz)^{\beta} - 1 \right) d \bX   \\
&= \frac{1}{N} \sum_{i=1}^N  \left( p_{\theta}(\bx_i | \bz)^{\beta} - 1 \right)
\end{split}
\end{equation}
The second integral can be written as:
\begin{equation}
\begin{split}
\int p_{\theta}(\bX | & \bz)^{\beta + 1} dX = \\
& \int \prod_{k=1}^K p_{\theta}( \bX = k \mid \bz)^{\beta + 1} \delta (\bX, k) d \bX  \\
&= \sum_{k=1}^K p_{\theta}(\bX = k \mid \bz)^{\beta + 1}
\end{split}
\end{equation}
Let's define $\beta$-cross-entropy for a single point for a categorical variable as:
\begin{equation}
\begin{split}
\mathbb{H}_{\beta}^{i-\textrm{cat}} &= - \frac{\beta + 1}{\beta} \frac{1}{N} \left( p_{\theta}(\bx_i \mid \bz)^{\beta} - 1 \right) \\
& \quad \quad + \frac{1}{N} \sum_{k=1}^K p_{\theta}(\bX = k \mid \bz)^{\beta + 1}.
\end{split}
\end{equation}
Then, the reconstruction loss for a single categorical sample using $\beta$-divergence can be written as
\begin{equation}
\mathcal{L}_{\textrm{REC-} \beta}^i = N \mathbb{E}_{\bz \sim q_{\phi}(\bz \mid \bx)} \left [ \mathbb{H}_{\beta}^{i-\textrm{cat}} \right ].
\label{eqn:L_rec_beta_cat}
\end{equation}
We can use the formulation derived in \citet{akrami2019robust} for continuous variables with the assumption of Gaussian distribution for $p(\bx_i \mid \bz) = \mathcal{N}(\bx_i, \sigma)$ which leads to
\begin{equation}
\begin{split}
\mathbb{H} & _{\beta}^{i-\textrm{continous}} = \\
&- \frac{\beta + 1}{\beta} \frac{1}{N} \left( \frac{1}{(2 \pi \sigma^2)^{\beta/2}} \exp \left( - \frac{\beta}{2 \sigma^2} \| \hat{\bx}_i - \bx_i \|^2 \right) - 1 \right)
\end{split}
\end{equation}
where $\hat{\bx}_i$ is the output of the decoder and the reconstruction loss becomes
\begin{equation}
\mathcal{L}_{\textrm{REC-} \beta}^i = N \mathbb{E}_{\bz \sim q_{\phi}(\bz \mid \bx)} \left [ \mathbb{H}_{\beta}^{i-\textrm{continous}} \right ].
\label{eqn:L_rec_beta_continous}
\end{equation}

For mixed tabular data, we use either equation \ref{eqn:L_rec_beta_cat} or \ref{eqn:L_rec_beta_continous} depending on the type of the variable. We summarize the training of RTVAE in algorithm \ref{alg:rvae}.

\section{Experimental Results}
We compare the performance between regular VAE and our RTVAE by gradually contaminating the training dataset with more outliers to evaluate robustness. We use three benchmark datasets made available by the cyber security community: KDDCup 99, NSL-KDD and UNSW-NB15. The goal is to detect cyber attacks at the network level. All datasets are in tabular format with categorical and continuous columns. We measured the area under the receiver operating characteristic curve (AUC) as an evaluation metric. 

\textbf{KDDCup 99:} \cite{kdd99} is the dataset used for ``The Third Knowledge Discovery and Data Mining Tools'' competition. The task was to build an automated network intrusion detector that can distinguish between attacks and normal connections. There are 41 columns of which 8 of them are categorical. We use the complementary 
10 \% data for training and the labeled test data for testing.

\textbf{NSL-KDD:} \cite{nslkdd} is the refined version of KDDCup 99 to resolve some of the inherent problems in KDDCup 99 dataset. More specifically, the redundant connection records were removed to prevent detection models become biased towards frequent connection records. We used the available full training dataset for training and test dataset for testing.

\textbf{UNSW-NB15:} \cite{unsw} dataset was introduced by a cyber security research team from the Australian Centre for Cyber Security. We used the available partitioned datasets for training and testing. The data has 43 columns out of which, 9 features are categorical. 

\textbf{Implementation Details:} We use fully-connected neural networks both in encoder and decoder with \textrm{tanh} and \textrm{softmax} activation functions for continuous and categorical variables, respectively. We use Python 3.6 for implementation \cite{oliphant2007python} using the pen-source libraries PyTorch \cite{paszke2019pytorch}, scikit-learn \cite{pedregosa2011scikit},  and NumPy \cite{walt2011numpy}. We use Adam \cite{kingma2014adam} as an optimizer with learning rate $1e-3$ and bias correction parameters $0.5$ and $0.999$ for gradients and squared gradients, respectively. We vary the $\beta$ parameter from $1 e-5$ to $0.1$ in logarithmic scale. Model selection for $\beta$ and the early stopping was done based on the best AUC from the hold-out validation dataset (20 \% of the training dataset).

\textbf{Results:} The results in Figure \ref{fig:results} show that the performance of the vanilla VAE degrades significantly even with a small amount of contamination (1 \%). Our RTVAE, on the other hand, stays robust to the outliers in the training datasets.

\section{Conclusion}
We derived a formulation on how to use robust $\beta$ divergence in a VAE framework for tabular datasets consisting of categorical and continuous features. Our results demonstrate that additional care needs to be taken when training with contaminated datasets with outliers. The RTVAE described here provides robustness with categorical data as shown in Figure \ref{fig:results}. 
%Although, we use intrusion detection as an application, our work can be used in other anomaly detection applications using tabular datasets.
\iffalse
\section*{Acknowledgements}
This work is supported by the following grants: R01-NS074980, W81XWH-18-1-0614, R01-NS089212, and R01-EB026299.
\fi
\nocite{langley00}

\bibliography{mybib.bib}

\begin{thebibliography}{22}
\providecommand{\natexlab}[1]{#1}
\providecommand{\url}[1]{\texttt{#1}}
\expandafter\ifx\csname urlstyle\endcsname\relax
  \providecommand{\doi}[1]{doi: #1}\else
  \providecommand{\doi}{doi: \begingroup \urlstyle{rm}\Url}\fi

\bibitem[ACCS()]{unsw}
ACCS.
\newblock Unsw-nb15.
\newblock
  \url{https://www.unsw.adfa.edu.au/unsw-canberra-cyber/cybersecurity/ADFA-NB15-Datasets/}.
\newblock [Online; accessed May 2020].

\bibitem[Akrami et~al.(2019)Akrami, Joshi, Li, Aydore, and
  Leahy]{akrami2019robust}
Akrami, H., Joshi, A.~A., Li, J., Aydore, S., and Leahy, R.~M.
\newblock Robust variational autoencoder.
\newblock \emph{arXiv preprint arXiv:1905.09961}, 2019.

\bibitem[An \& Cho(2015)An and Cho]{an2015variational}
An, J. and Cho, S.
\newblock Variational autoencoder based anomaly detection using reconstruction
  probability.
\newblock \emph{Special Lecture on IE}, 2\penalty0 (1), 2015.

\bibitem[Archive(1999)]{kdd99}
Archive, T. U.~K.
\newblock Kdd cup 1999 data.
\newblock \url{http://kdd.ics.uci.edu/databases/kddcup99/kddcup99.html}, 1999.
\newblock [Online; accessed May 2020].

\bibitem[Basu et~al.(1998)Basu, Harris, Hjort, and Jones]{basu1998robust}
Basu, A., Harris, I.~R., Hjort, N.~L., and Jones, M.
\newblock Robust and efficient estimation by minimising a density power
  divergence.
\newblock \emph{Biometrika}, 85\penalty0 (3):\penalty0 549--559, 1998.

\bibitem[Chandola et~al.(2007)Chandola, Banerjee, and
  Kumar]{chandola2007outlier}
Chandola, V., Banerjee, A., and Kumar, V.
\newblock Outlier detection: A survey.
\newblock \emph{ACM Computing Surveys}, 14:\penalty0 15, 2007.

\bibitem[Eduardo et~al.(2019)Eduardo, Naz{\'a}bal, Williams, and
  Sutton]{eduardo2019robust}
Eduardo, S., Naz{\'a}bal, A., Williams, C.~K., and Sutton, C.
\newblock Robust variational autoencoders for outlier detection in mixed-type
  data.
\newblock \emph{arXiv preprint arXiv:1907.06671}, 2019.

\bibitem[Eguchi \& Kato(2010)Eguchi and Kato]{eguchi2010entropy}
Eguchi, S. and Kato, S.
\newblock Entropy and divergence associated with power function and the
  statistical application.
\newblock \emph{Entropy}, 12\penalty0 (2):\penalty0 262--274, 2010.

\bibitem[Erfani et~al.(2016)Erfani, Rajasegarar, Karunasekera, and
  Leckie]{erfani2016high}
Erfani, S.~M., Rajasegarar, S., Karunasekera, S., and Leckie, C.
\newblock High-dimensional and large-scale anomaly detection using a linear
  one-class svm with deep learning.
\newblock \emph{Pattern Recognition}, 58:\penalty0 121--134, 2016.

\bibitem[for Cybersecurity()]{nslkdd}
for Cybersecurity, C.~I.
\newblock Nsl-kdd dataset.
\newblock \url{https://www.unb.ca/cic/datasets/nsl.html}.
\newblock [Online; accessed May 2020].

\bibitem[Futami et~al.(2017)Futami, Sato, and Sugiyama]{futami2017variational}
Futami, F., Sato, I., and Sugiyama, M.
\newblock Variational inference based on robust divergences.
\newblock \emph{arXiv preprint arXiv:1710.06595}, 2017.

\bibitem[Ghosh et~al.(2019)Ghosh, Sajjadi, Vergari, Black, and
  Sch{\"o}lkopf]{ghosh2019variational}
Ghosh, P., Sajjadi, M.~S., Vergari, A., Black, M., and Sch{\"o}lkopf, B.
\newblock From variational to deterministic autoencoders.
\newblock \emph{arXiv preprint arXiv:1903.12436}, 2019.

\bibitem[Hoffman \& Johnson(2016)Hoffman and Johnson]{hoffman2016elbo}
Hoffman, M.~D. and Johnson, M.~J.
\newblock Elbo surgery: yet another way to carve up the variational evidence
  lower bound.
\newblock In \emph{Workshop in Advances in Approximate Bayesian Inference,
  NIPS}, volume~1, pp.\ ~2, 2016.

\bibitem[Kingma \& Ba(2014)Kingma and Ba]{kingma2014adam}
Kingma, D.~P. and Ba, J.
\newblock Adam: A method for stochastic optimization.
\newblock \emph{arXiv preprint arXiv:1412.6980}, 2014.

\bibitem[Kingma \& Welling(2013)Kingma and Welling]{kingma2013auto}
Kingma, D.~P. and Welling, M.
\newblock Auto-encoding variational bayes.
\newblock \emph{arXiv preprint arXiv:1312.6114}, 2013.

\bibitem[M{\"u}nz et~al.(2007)M{\"u}nz, Li, and Carle]{munz2007traffic}
M{\"u}nz, G., Li, S., and Carle, G.
\newblock Traffic anomaly detection using k-means clustering.
\newblock In \emph{GI/ITG Workshop MMBnet}, pp.\  13--14, 2007.

\bibitem[Oliphant(2007)]{oliphant2007python}
Oliphant, T.~E.
\newblock Python for scientific computing.
\newblock \emph{Computing in Science \& Engineering}, 9\penalty0 (3), 2007.

\bibitem[Paszke et~al.(2019)Paszke, Gross, Massa, Lerer, Bradbury, Chanan,
  Killeen, Lin, Gimelshein, Antiga, et~al.]{paszke2019pytorch}
Paszke, A., Gross, S., Massa, F., Lerer, A., Bradbury, J., Chanan, G., Killeen,
  T., Lin, Z., Gimelshein, N., Antiga, L., et~al.
\newblock Pytorch: An imperative style, high-performance deep learning library.
\newblock In \emph{Advances in Neural Information Processing Systems}, pp.\
  8024--8035, 2019.

\bibitem[Pedregosa et~al.(2011)Pedregosa, Varoquaux, Gramfort, Michel, Thirion,
  Grisel, Blondel, Prettenhofer, Weiss, Dubourg, et~al.]{pedregosa2011scikit}
Pedregosa, F., Varoquaux, G., Gramfort, A., Michel, V., Thirion, B., Grisel,
  O., Blondel, M., Prettenhofer, P., Weiss, R., Dubourg, V., et~al.
\newblock Scikit-learn: Machine learning in python.
\newblock \emph{Journal of machine learning research}, 12\penalty0
  (Oct):\penalty0 2825--2830, 2011.

\bibitem[Walt et~al.(2011)Walt, Colbert, and Varoquaux]{walt2011numpy}
Walt, S. v.~d., Colbert, S.~C., and Varoquaux, G.
\newblock The numpy array: a structure for efficient numerical computation.
\newblock \emph{Computing in Science \& Engineering}, 13\penalty0 (2):\penalty0
  22--30, 2011.

\bibitem[Yao et~al.(2019)Yao, Liu, Zhang, and Peng]{yao2019unsupervised}
Yao, R., Liu, C., Zhang, L., and Peng, P.
\newblock Unsupervised anomaly detection using variational auto-encoder based
  feature extraction.
\newblock In \emph{2019 IEEE International Conference on Prognostics and Health
  Management (ICPHM)}, pp.\  1--7. IEEE, 2019.

\bibitem[Zhai et~al.(2017)Zhai, Chen, Zhang, and Wang]{zhai2017robust}
Zhai, Y., Chen, B., Zhang, H., and Wang, Z.
\newblock Robust variational auto-encoder for radar hrrp target recognition.
\newblock In \emph{International Conference on Intelligent Science and Big Data
  Engineering}, pp.\  356--367. Springer, 2017.

\end{thebibliography}
\bibliographystyle{icml2020}

\end{document}